\documentclass{article}

\usepackage{microtype}
\usepackage{graphicx}
\usepackage{subfigure}
\usepackage{booktabs} 
\usepackage{multirow}
\usepackage{mathtools}
\usepackage{array}
\usepackage{tablefootnote}

\usepackage{hyperref}

\usepackage[accepted]{mlsys2024}

\mlsystitlerunning{Shrinking the Giant : Quasi-Weightless Transformers for Low Energy Inference}

\begin{document}

\twocolumn[
\mlsystitle{Shrinking the Giant:\\ Quasi-Weightless Transformers for Low Energy Inference}




\begin{mlsysauthorlist}
\mlsysauthor{Shashank Nag}{ut}
\mlsysauthor{Alan T. L. Bacellar}{ut}
\mlsysauthor{Zachary Susskind}{ut/nvidia}
\mlsysauthor{Anshul Jha}{utsa}
\mlsysauthor{Logan Liberty}{ut}
\mlsysauthor{Aishwarya Sivakumar}{ut}
\mlsysauthor{Eugene John}{utsa}
\mlsysauthor{Krishnan Kailas}{ibm}
\mlsysauthor{Priscila M. V. Lima}{rio}
\mlsysauthor{Neeraja Yadwadkar}{ut}
\mlsysauthor{Felipe M. G. Franca}{felipe}
\mlsysauthor{Lizy K. John}{ut}

\end{mlsysauthorlist}

\mlsysaffiliation{ut}{Chandra Family Department of Electrical and Computer Engineering, The University of Texas at Austin, USA}
\mlsysaffiliation{utsa}{The University of Texas at San Antonio, USA}
\mlsysaffiliation{ibm}{IBM T.J. Watson Research Center, New York, USA}
\mlsysaffiliation{felipe}{Universidade do Porto, Instituto de Telecomunicações, Portugal}
\mlsysaffiliation{ut/nvidia}{UT Austin, USA *now at NVIDIA Research}
\mlsysaffiliation{rio}{Federal University of Rio de Janeiro, Brazil}
\mlsyscorrespondingauthor{Shashank Nag}{shashanknag@utexas.edu}

\mlsyskeywords{Transformers, Energy Efficient Inference, Multiplication-free, Accelerator, FPGA, ASIC}

\vskip 0.3in

\begin{abstract}

Transformers are set to become ubiquitous 
with applications ranging from chatbots and educational assistants
to visual recognition and remote sensing. However, their increasing computational and memory demands  is resulting in growing energy consumption. Building models with fast and energy-efficient inference is imperative to enable a variety of transformer-based applications. Look Up Table (LUT) based Weightless Neural Networks are faster than the conventional neural networks as their inference only involves a few lookup operations. Recently, an approach for learning LUT networks directly via an Extended Finite Difference method was proposed. We build on this idea, extending it for performing the functions of the Multi Layer Perceptron (MLP) layers in transformer models and integrating them with transformers to propose Quasi Weightless Transformers (QuWeiT). This allows for a computational and energy-efficient inference solution for transformer-based models. On I-ViT-T, we achieve a comparable accuracy of 95.64\% on CIFAR-10 dataset while replacing approximately 55\% of all the multiplications in the entire model and achieving a $2.2 \times$ energy efficiency. We also observe similar savings in experiments with the nanoGPT framework. 

\end{abstract}]

\printAffiliationsAndNotice{} 

\section{Introduction}\label{sec:intro}
Transformer based models have been of growing interest in recent times due to their versatile nature. With the advent of models such as ChatGPT, Gemini, and DALL-E, 
transformer-based generative models have assumed a greater significance than ever before, with applications across the fields of education, customer service, software development and healthcare. 
 In addition to excelling on language tasks, transformer based models have been demonstrating leadership positions even in vision tasks as illustrated by Vision Transformers in the Imagenet leaderboard \cite{imagenet-leaderboard}.

\begin{figure*}[htbp!]
  \centering
  {\includegraphics[width=0.9\textwidth, keepaspectratio]{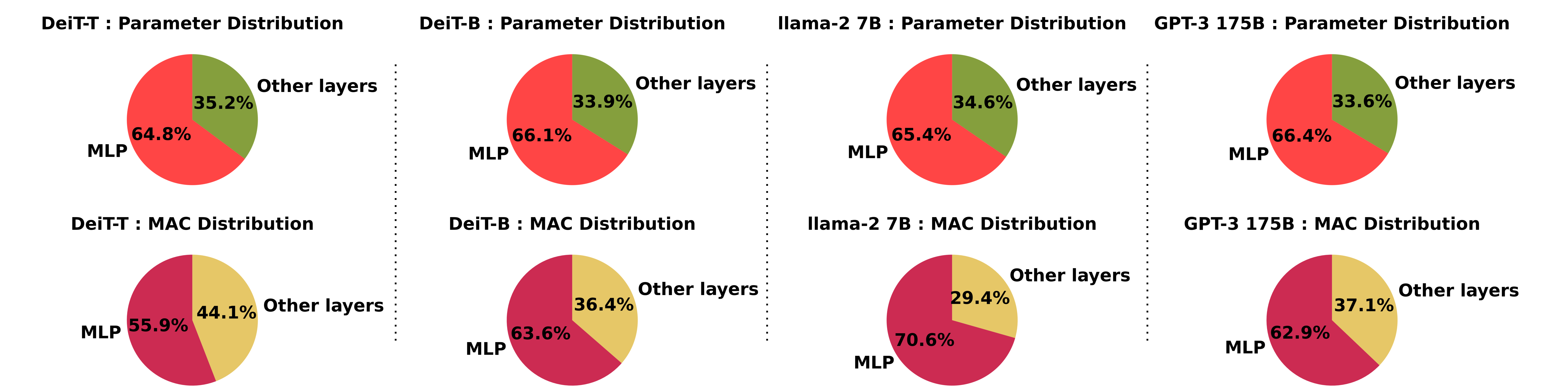}}
  \caption{Distribution of model parameters and MAC operations between the MLP and other layers in common Transformer-based models. The MLP layers contribute over 60\% of the overall model weights and about 50-70\% of the overall MAC operations.}
  \label{fig:intro_pie}
\end{figure*}

Although the aforementioned models have had tremendous capability and accuracy, their ever increasing model sizes pose significant challenges in their training and deployment. 
Over the years, the hardware required to train and deploy these models has not been able to keep pace with the model sizes, with transformer sizes growing at about 410 $\times$ every 2 years, as opposed to AI hardware memory growing at about 2$\times$ in the same interval \cite{aiwall-micro}. 
Notwithstanding the training costs, this issue of AI and memory wall makes inference deployment of these transformer-based models a real challenge. 
Owing to the large model sizes, these have high compute and memory requirements, resulting in a large energy consumption for each inference.
In typical real-world deployments these models serve a huge amount of inference queries - it is projected that the annual carbon emissions for inference servings of ChatGPT is 25$\times$ that of training, leading to 
growing concerns about their sustainability \cite{emission}. 
Characterization of transformer models reveals some interesting opportunities for reducing the energy consumption of these models.
We observe that in typical transformer models, a major fraction of compute and model weights is in the Multi-Layer Perceptron (MLP) layers of 
the model, as shown in Fig. \ref{fig:intro_pie}, based on the model configurations specified by OpenAI~\cite{openai-white}. While various model optimization techniques with transformers have been explored in the past, the core MLP block has continued to dominate the overall model, and it has been shown that these MLPs are infact the key behind the knowledge learnt in these models \cite{mlp-knowledge2}. 

The large fraction of multiply-accumulate (MAC) operations in the MLP layers indicate that making the MLP layers of the transformers energy-efficient would result in significant overall improvement in energy-efficiency.
We see an opportunity here by tapping into Weightless Neural Networks (WNNs), that are a class of Look-Up-Table (LUT)-based 
neural networks designed specially to be energy-efficient for inference, with the design matching the underlying logic fabric on hardware devices such as FPGAs and ASICs. These WNN models have demonstrated considerable performance improvements over MLP models, offering significant latency and energy advantages for iso-accuracy models  \cite{bthowen, uleen, dwn}.
Differentiable Weightless Neural Networks (DWNs) \cite{dwn} are a recent advancement that extends weightless models into MLP-like differentiable multi-layer architectures, achieving up to a 286$\times$ reduction in energy costs and a 60$\times$ reduction in latency compared to FINN \cite{finn}, the state-of-the-art Binary Neural Network (BNN) framework, under iso-accuracy scenarios. However, DWNs perform well only on simpler tasks and datasets, as they are unable to learn position-independent features.


\begin{figure}[htbp!]
  \centering
  {\includegraphics[width=0.4\textwidth, keepaspectratio]{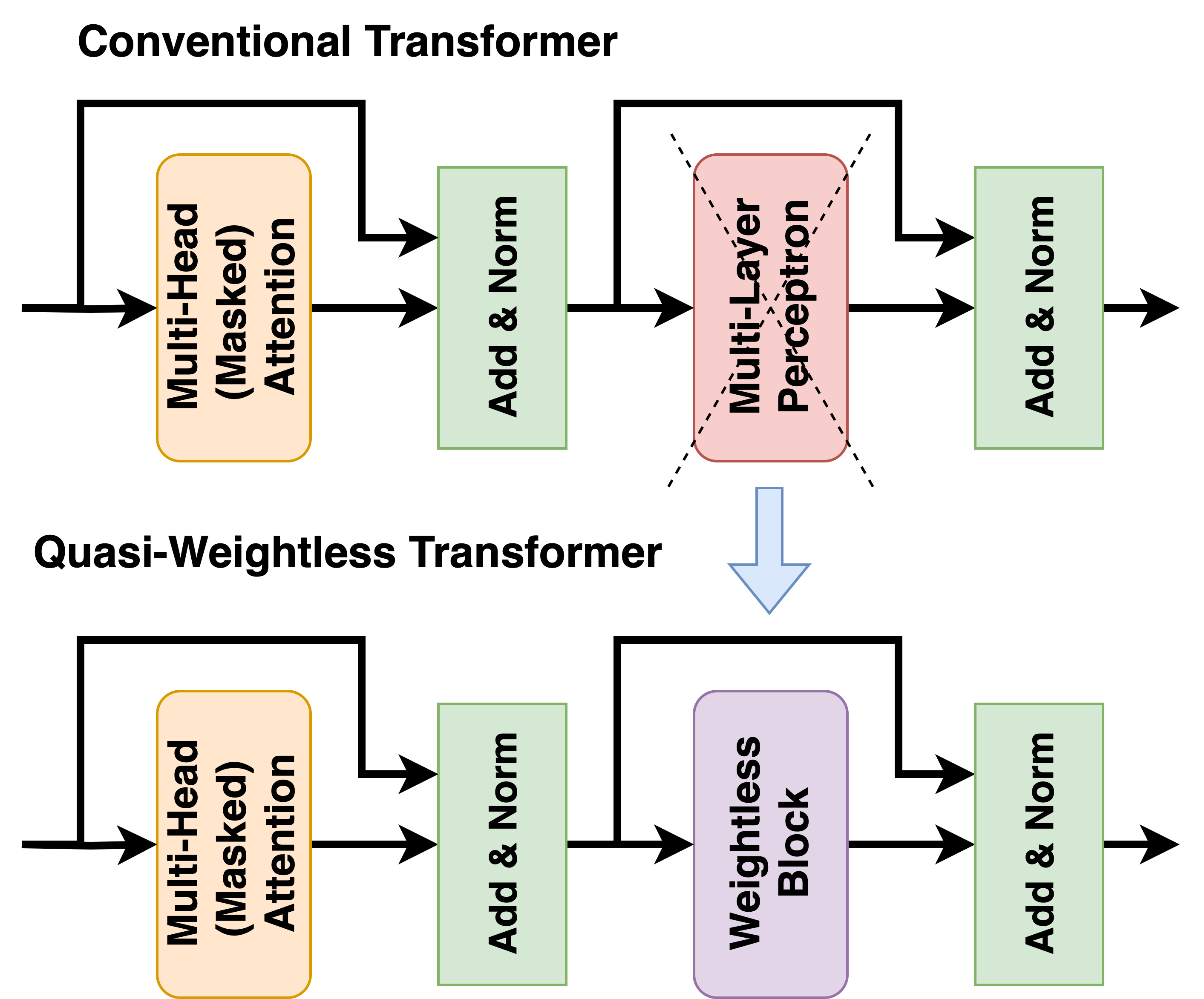}}
  \caption{Proposed Quasi-Weightless Transformer Design - overview of a single encoder/ decoder layer in the model}
  \label{fig:intro}
\end{figure}
In this work, we identify key features of transformers and WNNs that suggest combining aspects of these, and propose \textbf{QuWeiT} : \underline{Qu}asi-\underline{Wei}ghtless \underline{T}ransformers- a class of models that incorporate DWN layers in a transformer, completely eliminating the MLPs in them (Fig. \ref{fig:intro}). In doing so, we offer an inference-efficient solution that delivers the model performance advantages of transformers, while offering hardware performance benefits of WNNs. 
Specifically, we make the following contributions in this paper:
\begin{itemize}
     \item We propose a weightless block 
     incorporating DWN layers with modifications to the final layer and introducing a new conditional summation layer to enable seamless integration within large Deep Neural Networks (DNNs).
     \vspace{-2mm}
    \item We integrate the designed weightless block with encoder-based and decoder-based transformer models, and evaluate the model performance of these Quasi-Weightless Transformers (QuWeiT) on vision and language tasks. With this, for the first time, we demonstrate the versatility of weightless neural networks in shrinking large networks. 
    \vspace{-2mm}
    \item We propose an accelerator design for QuWeiT, and evaluate its performance on FPGA and ASIC against baseline accelerator designs.
\end{itemize}
\vspace{-2mm}

We note that low energy requirements, ultra-low latency and a LUT based implementation of the weightless layers makes QuWeiTs excellent candidates for inference deployments. We demonstrate the efficacy of these models on a set of workloads, indicating this model design could be extended further to offer efficient lightweight energy-efficient alternatives to Large Language Models (LLMs).

\begin{figure*}[htbp!]
  \centering
  {\includegraphics[width=0.8\textwidth, keepaspectratio]{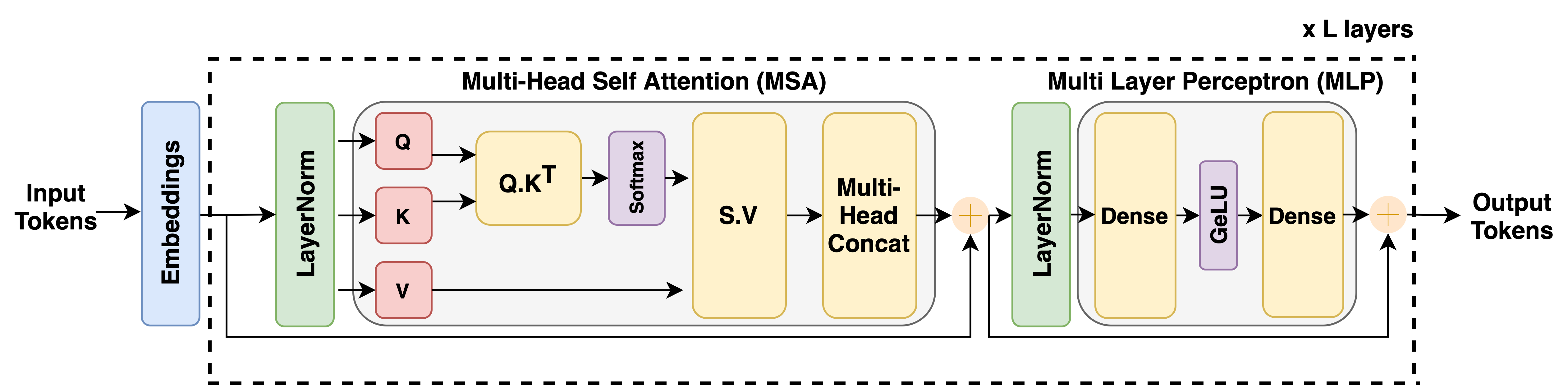}}
  \caption{A typical transformer model architecture with an encoder-only or decoder-only stack of layers. In the case of decoder-only models, the MSA shown here is a masked self-attention, with future tokens masked during attention score computation.}
  \label{fig:transformer}
\end{figure*}

\vspace{-2mm}
\section{Background and Motivation} \label{sec:background}
\subsection{Transformers and their model architecture}

Transformers are deep-learning models that are based on the principle of deriving context or relationships on a series of input tokens, and using it to predict a likely output token. These models have gained prominence for their ability to understand and generate natural language with remarkable accuracy and coherence on natural language processing (NLP) tasks. Transformers form the bedrock of most of the LLMs and foundational models, and are often trained on  
vast amounts of data. In the recent past, these models have also been extended to the realm of computer vision tasks, with different variants of Vision Transformers designed for common image classification tasks. Fig. \ref{fig:transformer} shows a typical transformer model architecture that takes an input prompt, 
generates query ($Q$), key ($K$) and value ($V$) matrices from it, applies self-attention, and passes it through the MLP layer to generate an output response, one token at a time. 

    

While conventional natural language processing tasks worked with RNNs \cite{rnn-nlp}, the advent of self-attention revolutionized the field \cite{vaswani}. 
Newer models have largely had the same transformer-based architecture, with many of them switching to a decoder-only architecture \cite{gpt, llama}. 
Amongst the recent works, GPT \cite{gpt}, Megatron \cite{megatron}, and LLAMA (Large Language Model Meta AI) \cite{llama} stand out as  advanced variants tailored specifically for handling complex language tasks. While these models are architecturally similar to the original decoder-based models, they are much larger in size, with several layers of decoders stacked to make up a few tens-hundreds of billion parameters. 

Similarly, on the computer vision front, several Vision Transformer (ViT) models have achieved state-of-the-art performance on image classification tasks \cite{vit,deit,vitg}. These models have
the same architecture as the regular transformer with an encoder-stack instead, and an additional class token is appended to determine the predicted class.





\begin{table*}[ht!]
\centering
\caption {Breakdown of computations and model weights in an encoder-only or decoder-only model. $n_{layers}$ represents the number of layers, $N$ represents the context length, and $D$ represents the latent dimensions of the model. In GPT-3 175B, $n_{layers}=96$, $N=4096$, $D=12288$. We consider the typical case where the MLP hidden layer dimensions are $4\times$ the model dimensions. 
}
\vskip 0.15in
\begin{tabular}{ ccccc }
 \toprule
 Layer & \# Parameters & \# MAC ops per token  & GPT-3 : \# Parameters & GPT-3 : \# MAC ops\\
 \midrule
 Q, K, V Projection & $3\times n_{layers} \times D \times D$ &  $3 \times n_{layers} \times D \times D$  & 43,486,543,872 & 43,486,543,872 \\
 $Q.K^T$ & - & $ n_{layers} \times D \times N$ & - & 4,831,838,208  \\
 SoftMax.V & - & $ n_{layers} \times N \times D$ & - & 4,831,838,208 \\
 Multi-head concat & $ n_{layers} \times D \times D$ & $n_{layers} \times D \times D$ & 14,495,514,624 & 14,495,514,624\\
 \midrule
 Feed-forward 1 (MLP) & $4 \times n_{layers} \times D \times D$  &  $4\times n_{layers} \times D \times D$ & 57,982,058,496 & 57,982,058,496\\
 Feed-forward 2 (MLP) & $4\times n_{layers} \times D \times D$  &  $4\times n_{layers} \times D \times D$ & 57,982,058,496 & 57,982,058,496 \\
 \bottomrule
\end{tabular}
\label{table:workload}
\end{table*}

\vspace{-1mm}
\subsection{Optimizing Transformer Models}

In order to address the issues with increasing model sizes of transformer-based models and the consequent deployment implications, various model pruning techniques have been studied in the past. Wanda \cite{prunellm} proposes a pruning technique to introduce sparsity in LLMs without requiring additional fine-tuning. LLM Pruner \cite{llmpruner} is another approach that performs structured pruning of Language Models. 

Similarly, quantization of such models has been investigated, which helps reduce the memory requirement and compute precision. SmoothQuant \cite{smoothquant}, Q8BERT \cite{q8bert} and \cite{ptqllm} have proposed different quantization techniques, including quantization-aware training, and post-training quantization upto 8-bits. In the realm of vision tasks, I-ViT \cite{ivit} and BinaryViT \cite{binaryvit} have explored quantized designs that are suitable for efficient inference implementations.


\vspace{-1mm}
\subsection{Workload Analysis and Key Insights}\label{sec:insights}

We study a range of transformer models to characterize the computations and memory usage in their constituent layers, and summarize the findings in Table \ref{table:workload}. For instance, analyzing the 96 layers of GPT3, and separating out the MACs and parameters in the various layers, we can learn identify the 
proportion of computes and weights in the various layers.
As illustrated here, and as alluded in Fig. \ref{fig:intro_pie}, particularly of interest is the fact that across typical transformer models, MLP layers contribute over 60\% of the total model weights. At the same time, these layers also contribute to a significant fraction of the overall compute - ranging between 50 - 70\% of the total MAC (multiply-accumulate) operations. These findings are also corroborated in \cite{vita}.

Having made these observations, we now turn our focus towards Weightless Neural Networks (WNNs), in an effort to identify potential energy-efficient alternatives to these MLP layers. 

\begin{figure}[htb!]
\centerline{\includegraphics[width = 7cm]{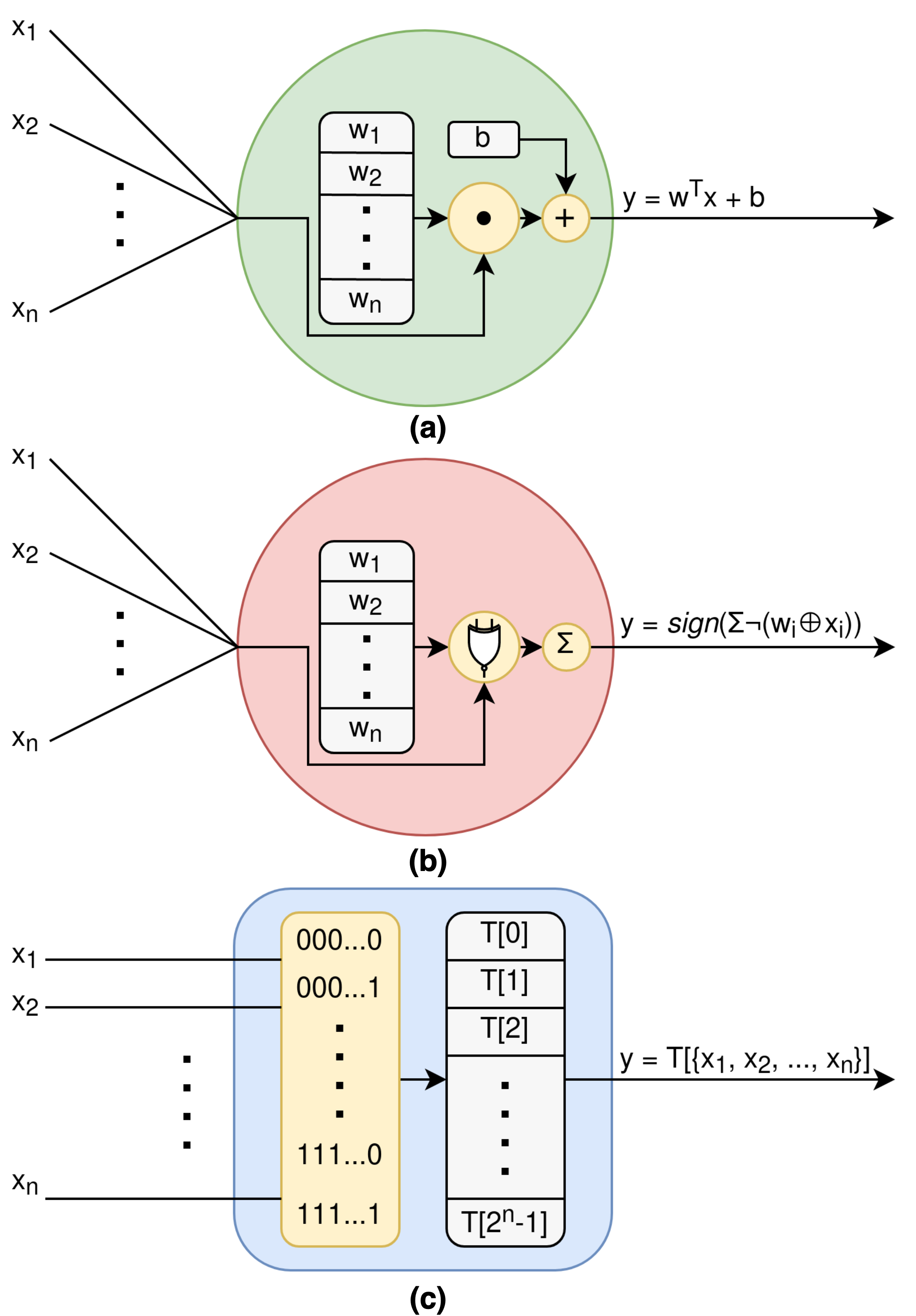}}
\caption{(a) Conventional Neuron : Each neuron multiplies inputs with weights and adds them. (b) Binary Neural Network Neuron : The weights being binary, the multiplication operation is substituted by a XNOR (c) Weightless Neuron : In contrast, the input sequence is concatenated and used to "look up" in the LUT with no MAC operations involved.}
\vspace{-4mm}
\label{fig:neuron}
\end{figure}

\vspace{-1mm}
\subsection{Weightless Neural Networks (WNNs)}


Weightless Neural Networks are neural networks inspired by the dendritic trees of biological neurons. 
These weightless networks are primarily comprised of look-up table (LUT) based neurons, and eliminate the need for power and resource hungry multiply-accumulate operations in conventional neurons (Fig. \ref{fig:neuron}). 
A $n$-input lookup table is a highly expressive structure which can represent any one of $2^{2^n}$ possible functions. Consequently, a LUT-based neuron has a higher learning capacity, with a VC dimension of $2^n$ \cite{vcwnn}, compared to a DNN neuron, which has a VC dimension of $n + 1$.
While earlier works in this field \cite{wisard,wisard_fpga} suffered from high memory requirements, recent works like BTHOWeN \cite{bthowen} and ULEEN \cite{uleen} 
have demonstrated significantly lower compute and memory requirements compared to iso-accuracy MLPs and CNNs for image classification tasks. This has sparked a renewed interest in WNN research and potential usage. Additionally, the weightless neural network has a significantly lower latency since only a small number of table lookups are involved in the inferencing. While software models of WNNs suffer from a higher training complexity on GPUs due to the intense bit-wise and sparse operations, they are well-suited for deployment on FPGAs (that have underlying LUT slices) or ASIC implementations. We also note that since we primarily aim to design an inference-efficient model, the training complexity is not a major concern. 
Differentiable Weightless Neural Networks \cite{dwn} is the most recent work that offers a class of multi-layer differentiable weightless neural networks, comprised of a network of Look-up-tables (LUTs). By introducing an Extended Finite Differentiation based approximate-differentiation technique for LUT entries and inputs, DWN enables a fully-differentiable multi-layer weightless solution. By further incorporating features such as Learnable Mapping and Learnable Reduction, DWN achieves outstanding performance on structured and tabular data. Futhermore, in iso-accuracy scenarios, DWN surpasses FINN \cite{finn} in terms of model size, latency, and energy efficiency, achieving a reduction of up to 2000x in area-delay product~\cite{dwn}. DWNs were also seen to significantly excel over recent works such as LogicNets~\cite{logicnets}, NeuraLUT~\cite{neurallut}, PolyLUT~\cite{polylut}, and DiffLogicNet~\cite{petersen2022}. However, these models perform poorly on image classification tasks such as the CIFAR-10 \cite{cifar10} dataset, as these do not implement an architecture that can learn positional independence of features. 




\subsection{Combining WNNs and Transformers}


As outlined earlier, transformers incur high resource utilization, high latency, and involve several power-hungry multiply-accumulate operations. On the contrary, WNNs offer a low-latency energy efficient solution involving look-up operations, resulting in low resource utilization. 
Futhermore, while transformer model sizes scale quadratically with latent dimensions, we demonstrate in Section \ref{sec:eval} that WNNs have the potential to scale linearly. 
By incorporating self-attention, transformers are able to learn positional independence across tokens and perform well on complex datasets; whereas the current WNNs struggle with positional invariance. Lastly, as illustrated in the previous sections, MLP layers in transformers are the most compute and memory intensive. 
Various studies in the past have demonstrated that MLP layers are essentially the knowledge-house of transformer models \cite{mlp-knowledge2, mlp-knowledge3}, encoding crucial knowledge observed from trained data within them. Prior research in the field of WNNs demonstrates that these networks can efficiently learn patterns represented by MLP layer. Additionally, prior studies indicate that WNN models can achieve similar model accuracy as MLP models on a range of tasks and datasets, while significantly outperforming those in terms of reduced latency and reduced energy consumption  \cite{dwn, uleen, bthowen}. 
Thus, all of these factors suggest that many of the aspects of transformers and WNNs are complementary to each other, and the possibility to combine these is worth studying. 

We note that transformer models have repeated blocks of a few layers stacked one over the other. 
As a consequence of the repeated blocks, tackling the MLP would involve modifying intermediate layers of a larger network. 
In this regard, we need to redesign and adapt weightless layers that preserve the dimensions of the latent space of transformer network before integrating them together. 




Based on these inferences, we propose to replace the MLP layers of transformer models with similarly designed fully connected (FC) weightless layers to come up with Quasi-Weightless Transformers. We note that this is a non-trivial integration owing to the challenge discussed above. WNNs have traditionally only been used to design small models for simple tasks, and such an adaptation into a giant model has not been explored in the past, to the best of our knowledge. This change would be particularly relevant if it can be achieved without accuracy drops.
By doing so, we could leverage WNN's hardware and energy savings advantages combined with the ability to learn positional invariance through the self-attention layers of the transformers. We also note that since the LUTs can inherently learn non-linearities, we would not require additional non-linear activation layers that would be otherwise required within an MLP layer.

\section{Quasi-Weightless Transformers} \label{sec:design}


\subsection{Differentiable Weightless Block}\label{sec:layerdesign}

\begin{figure}[htb!]
\centering
 {\includegraphics[width=0.45\textwidth, keepaspectratio]{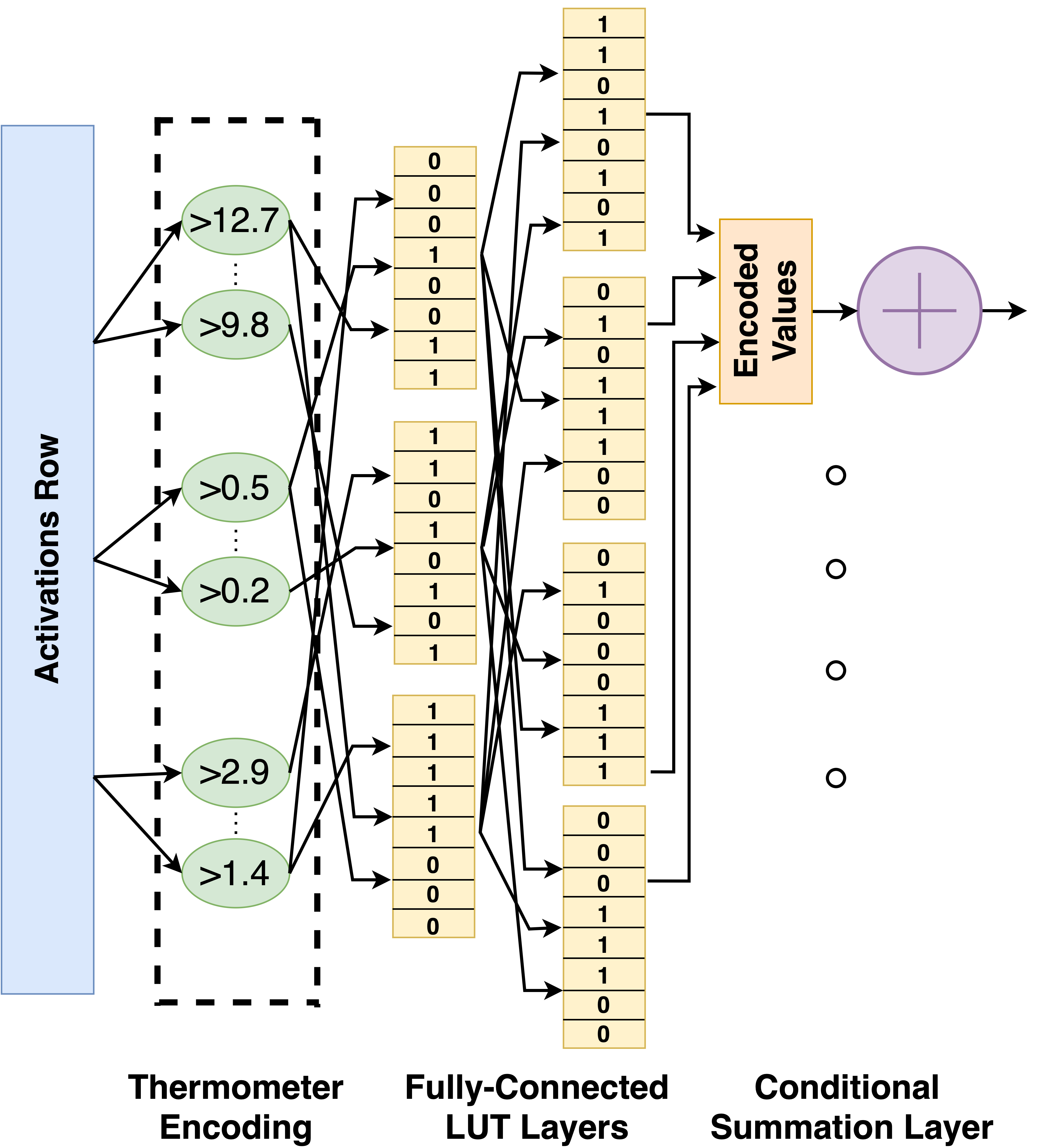}}
 \caption{Differentiable Weightless Block that replaces the MLP} 
\vspace{-4mm}

 \label{fig:weightless_block}
\end{figure}

{As discussed, DWN \cite{dwn} introduces fully differentiable weightless layers. While the original models target classification tasks and employ binary values in the final LUT layer for optimal efficiency, adapting DWNs as intermediary blocks within traditional DNN layers requires real valued outputs. To address this, we propose using a conditional summation layer following the final layer of LUTs, that adds higher precision (floating point) encoded values based on the output requirements from the block. The conditional summation layer operates as follows: if the output from the LUT in the last layer is a $1$, an encoded value is added to the corresponding output, and if it is $0$, it is simply skipped. 
Importantly, this layer incurs no additional multiplication overhead during inference, as it merely adds an encoded value based on which LUTs participate in the summation - while still maintaining the full-precision required by the model. 

Figure \ref{fig:weightless_block} illustrates the proposed differentiable weightless block.
To emulate the independent processing of tokens typical to MLP layers, we flatten input activations row-wise and pass them through these layers one row at a time. After a thermometer encoding layer~\cite{thermometer, dwn}, which converts the input activations into a sequence of bit representations, we utilize one or more configurable LUT layers, ending with the conditional summation layer. The output layer contains a number of summation units matching the transformer's latent dimension $D$, preserving the network’s latent dimensionality. We note that for smoother gradients during training, these encoded values use fp32 precision. However, post-training these values can be quantized to a lower precision as required by the rest of the network.}

\vspace{-2mm}
\subsection{Integrated Quasi-Weightless Transformer Design}

\begin{figure}[htbp!]
  \centering
  {\includegraphics[width=0.5\textwidth, keepaspectratio]{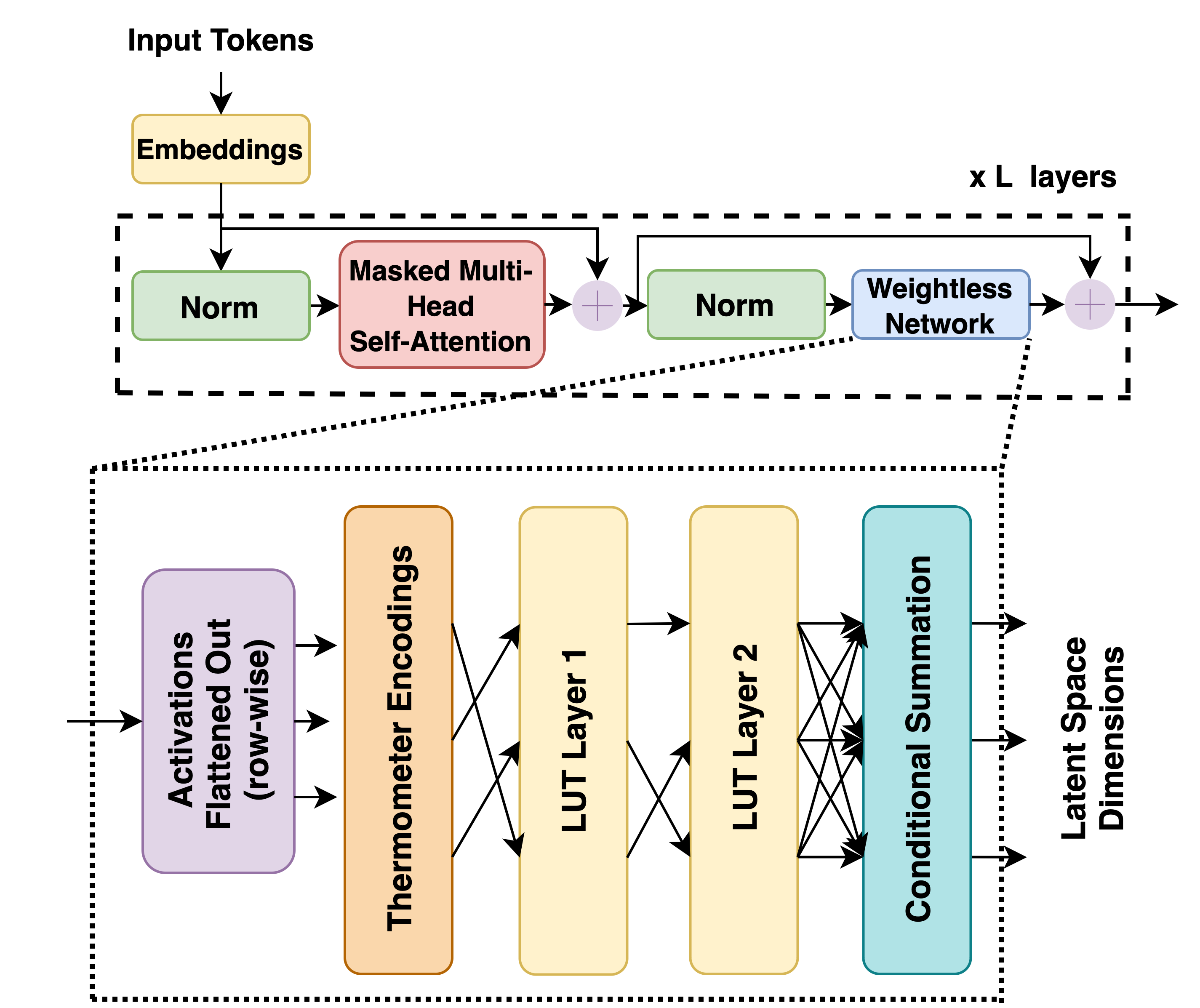}}

  \caption{Proposed Quasi-Weightless Transformer with the integrated Differentiable Weightless Block}
  \vspace{-2mm}
  \label{fig:sys_design}
\end{figure}

Fig. \ref{fig:sys_design} shows the proposed design of the Quasi-Weightless Transformer. We integrate the differentiable weightless block into each of the encoder/ decoder units in the stack, with the original MLP layer being replaced by it. By using DWN layers, the gradients are well-defined for the LUT inputs and entries, and the entire quasi-weightless transformer model can be trained end-to-end. 

\subsection{Processing Element Block Design}

We propose a dedicated processing element (PE) block design for the weightless block, to develop an optimized accelerator design for QuWeiT model inference. As this work focused on developing these as inference-efficient models, the accelerator proposed is primarily an inference-only accelerator design.  
We provide a detailed description of its implementation, tailored for both FPGA and ASIC.

\begin{figure}[htbp!]
  \centering
  {\includegraphics[width=0.45\textwidth, keepaspectratio]{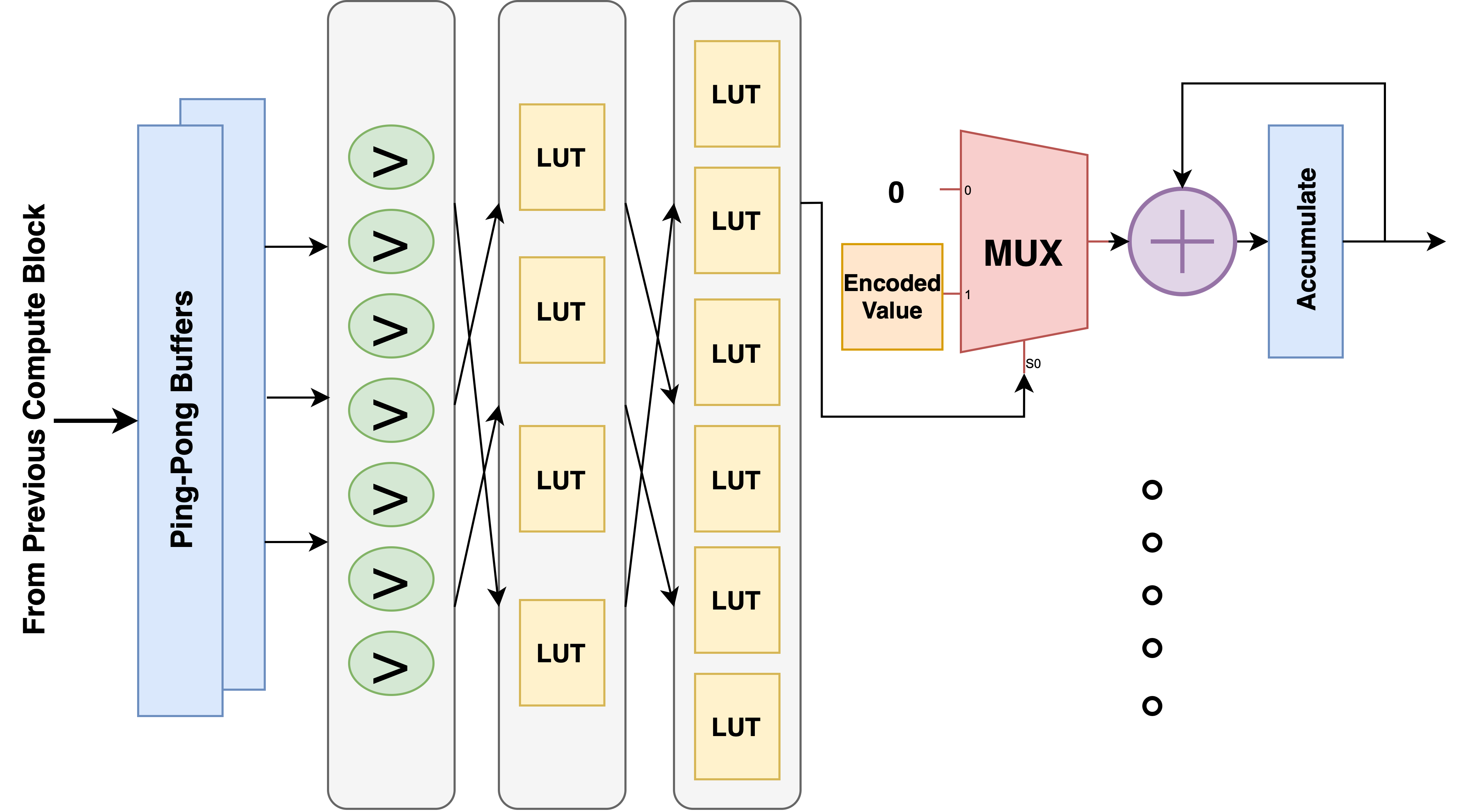}}
  \caption{Processing Element Design for the Weightless Block}
  \vspace{-1mm}
  \label{fig:hw_design}
\end{figure}

We design our PE block based off the implementation proposed in DWN \cite{dwn}, as shown in Fig. \ref{fig:hw_design}. Within the weightless block, all the output bits are required to infer input addresses to the next layer of LUT based on the connections. Moreover, as all the LUTs would have to be implemented on hardware uniquely, we process all the elements parallely. 
Consequently, all elements in a row of the input activations to the weightless block are processed together, with consecutive rows processed sequentially. Ping-pong buffers at the input of the PE enable the activations from the previous layer to be buffered in, while also allowing another set of activations to be processed through the weightless block. For the thermometer encoding and fully-connected LUT layers, we utilize the same implementation as in DWN - with the LUT neurons being mapped onto the LUTs within the Configurable Logic Blocks (CLBs) of FPGA, or as gates of ASICs. 

For the conditional summation layer, we introduce an adder circuit preceded by a 2:1 MUX for each dimension in the latent space. While we could also introduce an adder tree for each of these for low-latency, this would increase the resource consumption quite significantly. We note that there are additional considerations we need to take. Particularly, these weightless blocks are intermediate blocks in a larger neural network. Therefore, the overall throughput of the network is determined by the slowest layers. Since we are not optimizing the attention and norm layers, those would quite likely be the bottleneck for the overall model throughput. Hence, we introduce registers and accumulate the output activations for the weightless block over multiple cycles, roughly time-matching with the preceding and following self-attention layers. For the non-weightless layers in the network, we use a regular systolic array based accelerator implementation as used in literature typically \cite{mlpsystolic}, and stream these inputs to the weightless block component on the same device. 

An important aspect to note here is that by introducing weightless layers instead of MLP layers, we are not trading off compute with additional memory. In our target devices, FPGAs and ASICs, compute blocks are majorly built up of LUTs and gates respectively. Consequently any compute or processing element block design for conventional neural network acceleration get mapped to these logic elements on device. In of our proposed model, the LUTs in the weightless layers would get directly mapped to these LUTs and gates on device - incurring no additional overhead on the memory. Infact on the contrary, the LUT-based implementation eliminates the memory requirement associated with those weights that would have been present in a conventional layer - while also eliminating the associated off-chip memory access. 

\section{Evaluation} \label{sec:eval}
\subsection{Experimental Methodology}

To evaluate and iterate upon our design, we adopt the following methodology : we setup a baseline model, and design the corresponding QuWeiT model replacing MLPs with the configurable differentiable weightless block. We train this model, evaluate its performance, generate RTL for the weightless block's PE and baselines, and go about the FPGA/ ASIC flow to get area consumption and energy reports. We then iterate over our design incorporating feedback from these empirical estimates to achieve an iso-accuracy model at a reasonable area-energy cost.

We define the weightless block in PyTorch, and integrate these with the baseline models we are interested in evaluating. We alter the model definition, and train these end-to-end on the dataset of interest using a similar learning rate, scheduler and optimizer on an A100 GPU. Loading pretrained weights for the non-weightless layers appeared to have minimal to no impact on model performance, and hence we let the model train from scratch in most cases. 

For each transformer model variant that we integrate with, we experiment with various training strategies, and configurations of the weightless block to achieve iso-accuracy models, and scale these hyperparameters roughly based on the size of MLP that is being eliminated.


We evaluate the efficacy of QuWeiT's performance on target hardware by estimating the performance of our proposed accelerator design, and comparing it against standard implementations for the baseline transformer models. In order to provide these end-to-end estimates, we write RTL code and synthesize designs for both the weightless block's PE, as well as the other conventional transformer layers, including the MLP block that we eliminate.

RTL code for the weightless block's PE is generated from the QuWeiT PyTorch model using custom mako templating \cite{mako}. A python script is used to parse the model checkpoint, and dump the thermometer encoding thresholds, LUT entries, summation encoded values, and mapping information for each weightless block in the network. The mako template scripts parse through these dumped model files, and create SystemVerilog codes instantiating LUTs with encoded values and interconnects. 


To setup baselines for comparison, and for the non-weightless layers in QuWeiT, we consider a pipelined systolic array based accelerator design, as is typically used in the literature for efficient blocked matrix multiplication implementations \cite{mlpsystolic,systolic-micro}. To provide additional points of comparison, including area-energy tradeoffs, we consider different dimensions of systolic arrays : 8$\times$8, 16$\times$16, 32$\times$32, and 64$\times$64. 
We describe further specifics of this while discussing hardware performance in Section \ref{sec:hwperf}. Parametrized RTL codes for these implementations were written in Verilog HDL. 


Looking back at the detailed transformer architecture in Fig. \ref{fig:transformer}, the key computations involved in a single sample inference of the transformer include a series of general matrix multiplication (GEMM) operations, specifically in the generation of $Q$, $K$, $V$ matrices, $QK^T$ computation, $Softmax.V$ computation, multi-head concatenation, and the two linear transformations within the MLP layer. Each of these computations involve a GEMM operation of two matrices of deterministic size. Based on the chosen systolic array configuration, we determine the number of cycles it takes to process each of these GEMMs, and project the corresponding energy consumed based on the power consumption of the design, as described subsequently.  




For evaluation on \textbf{FPGA}, we consider the Xilinx xcvu9p-flgb2104-2-i part as our target device. We use the Xilinx Vivado 2022.1 tool and synthesize the RTL with a target clock frequency of 200 MHz. The synthesized design is placed and routed on the target FPGA without utilizing DSPs, and hardware resource utilization reports are generated, indicating the LUT, FF, and Block RAM (BRAM) utilization. We then perform a vectorless power estimation for this design, using a default switching activity factor of 12.5\% to estimate the dynamic power in the reports.  

To perform an area and power analysis of our design on a custom \textbf{ASIC}, we employ the Synopsys Design Compiler Q-2019.12-SP5-5 logic synthesis tool. The generated RTL is parsed through to ensure syntactic correctness, and the RTL is elaborated into an intermediate format that captures the design hierarchy. Post this, the tool performs optimization and synthesis, mapping the RTL to the standard cells in the specified technology library. Technology libraries consist of information about the target process technology, including detailed descriptions of standard cell comprising of logic gates and flip flops. For our experiments, we use the 45nm process technologies, and synthesize at a target clock frequency of 1GHz. Following synthesis, the power optimization technique is employed to ease the dynamic power by clock gating and data path reformation. Area optimization is also performed to reduce the logic cell count and minimize the area, while meeting the power constraints. The final gate-level netlist, and detailed area and power reports are generated by the tool at the end, and we work with the total power reported.

Based on the these generated reports, we extrapolate the energy required per sample inference for both the conventional transformers and QuWeiT to arrive at empirical estimates. For the conventional transformer, we compute the total energy required to compute all the GEMMs through the systolic arrays, alongwith the energy required to compute the non-linear scalar operations (softmax, GeLU, and others), considering a steady state operation. For QuWeiT, we consider a system where the systolic array block is used to compute the self-attention layers that are retained from the conventional transformer, and the outputs from here are streamed into the weightless block's PE on the same device through the ping-pong buffers. Here, we project the energy consumption for the GEMMs corresponding to the self-attention block to be processed on the systolic array, and for the weightless block to be processed through its dedicated processing unit. 



\subsection{Case Studies}
\subsubsection{Vision Transformers}
\begin{figure}[htbp!]
  \centering
  \includegraphics[width=0.4\textwidth, keepaspectratio]{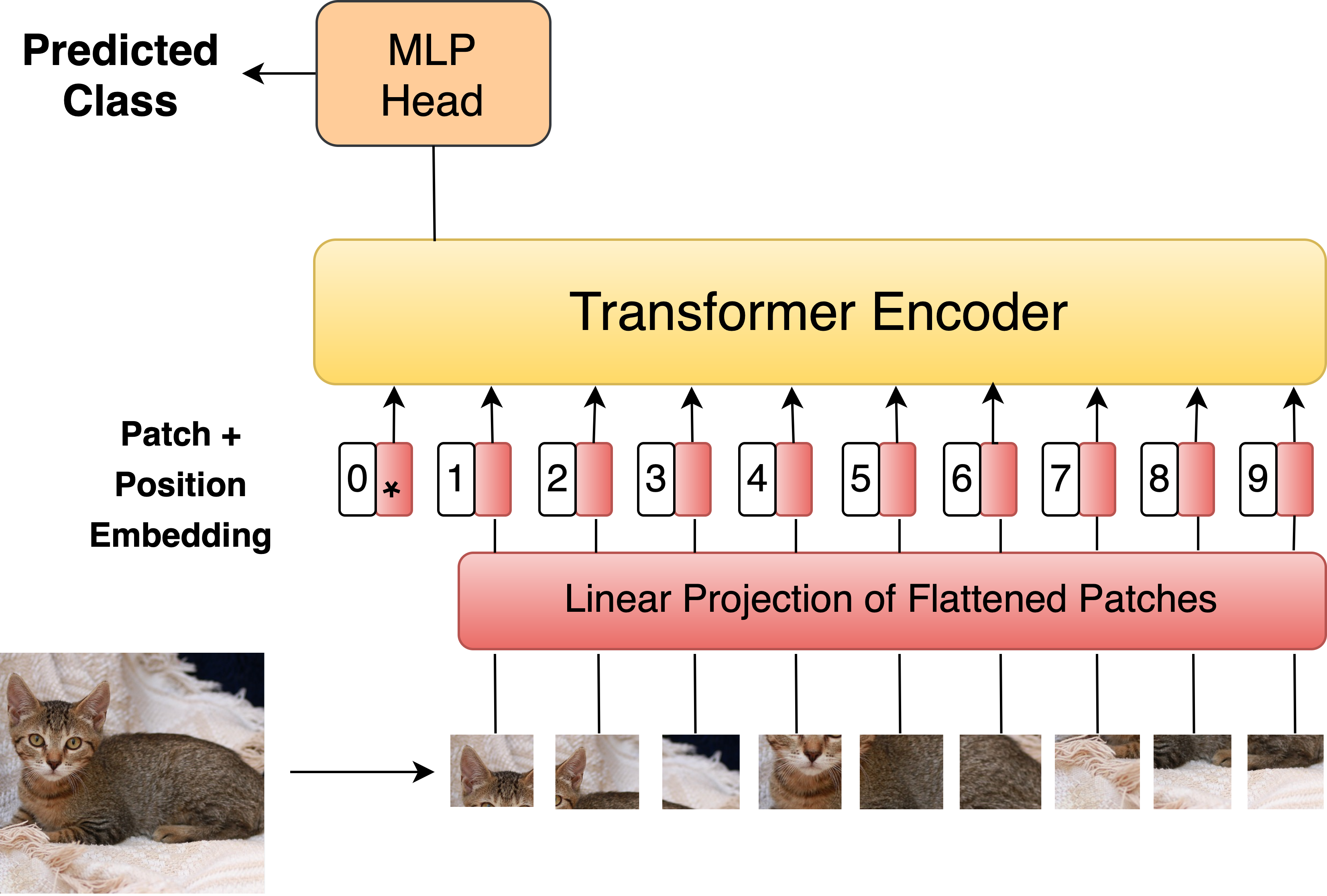}
  \caption{A typical vision transformer model consisting of a stack of encoder blocks. Figure adapted from \cite{vit}. 
  }
  \label{fig:vit}
\end{figure}

Vision Transformers are encoder-based models for computer vision tasks. As shown in Fig. \ref{fig:vit}, these involve self-attention applied to a series of tokens generated by splitting the image into patches. In case of an image classification task, a classification head is attached to the sequence, and extracted at the end of the encoder stack to get the predicted class. 
Prior studies have explored quantization techniques to make these models more efficient for hardware deployment \cite{vit-quant-nips,vit-quant-survey}. I-ViT \cite{ivit} proposes a fully int-8 quantized vision transformer variant, that is particularly suitable for inference on dedicated hardware, and we use this in our evaluation.  

\subsubsection{GPT-like Models}

GPT models are a set of Large Language Model (LLMs) used for generative AI applications, and in many of the current natural language processing (NLP) tasks. These models are comprised of decoder-only stack, and are architecturally similar to most of the latest LLMs \cite{opt,llama,gpt4}. 
We employ nanoGPT \cite{nanogpt}, a fast and efficient implementation of GPT models, to train a small GPT-like network and evaluate its model performance in terms of test loss. 

\subsection{Results}
\subsubsection{Model Accuracy}

We first evaluate our proposed QuWeiT technique in the context of vision transformers.
We consider the I-ViT-T model \cite{ivit} as our baseline, and design the corresponding QuWeiT with 8-bit thermometer encoding, two DWN layers with $768$ and $192$ LUTs in them respectively, and int8 encoded values. These parameters were picked to roughly scale linearly with the latent dimension, where the model dimension ($D$) and the hidden-layer dimensions ($4\times D$) were $192$ and $768$ respectively. We train this model and compare it against the baseline. 
As noted in Table \ref{table:acc}, with CIFAR-10, the QuWeiT model achieves an accuracy of 95.5\% against a baseline model accuracy of 94.8\% \cite{affine}. 
On the CIFAR-100 dataset as well, the QuWeiT model reports comparable accuracy of 78.8\% against a baseline accuracy of 79.2\%. 
For both these datasets, we use the standard data augmentation used in DeiT \cite{deit} and upscale the image to $224\times224$.

These findings suggest that the quasi-weightless models perform comparably to the baseline in terms of model accuracy. We also note that these weightless models that have been integrated with vision transformers perform much better compared to prior weightless neural networks, that report the accuracy on CIFAR-10 in the range of 50-60\% \cite{dwn}.

In the context of language tasks, we evaluated a baby-GPT model provided in the nanoGPT framework \cite{nanogpt} on the shakespeare-char dataset. The corresponding QuWeiT model designed with similar scaling and fp32 encoded values performs comparably as shown in Table \ref{table:acc}, with a loss of 1.465 against a baseline of 1.469. 

\begin{table}[htbp!]
\centering
\vspace{-3mm}
\caption {Model Accuracy / Loss comparisons of baseline transformer models vs. the designed QuWeiT models.\textsuperscript{*}\cite{ivit}, \textsuperscript{\textdagger}\cite{nanogpt} nanoGPT-baby. \textsuperscript{1}\cite{affine},  \textsuperscript{2} We train this baseline model based on the model repository, and verify by correctly replicate the baseline for CIFAR-10. 
}
\vskip 0.1in

\begin{tabular}{ p{2cm}p{1.75cm}p{1.3cm}p{1.3cm} }
 \toprule
 Model Variant & Dataset & Baseline & QuWeiT\\
 \midrule
 I-ViT-T\textsuperscript{*} & CIFAR-10  & 94.8\%\textsuperscript{1}  
 & 95.5\%\\
 I-ViT-T\textsuperscript{*} & CIFAR-100 & 79.2\%\textsuperscript{2}
 & 78.8\% \\
 nanoGPT
 \textsuperscript{\textdagger} & shakespeare
 & 1.469 & 1.465 \\ 
 \bottomrule
\end{tabular}
\label{table:acc}
\end{table}


\begin{table*}[t!]

\centering
\newcolumntype{C}[1]{{\centering}m{#1}}
\caption{Performance comparison of systolic array implementations of the GEMMs in MLP layer in I-ViT-T and the PE implementation of the weightless block in the corresponding QuWeiT. The MLP in this network involves two GEMMs of dimensions (196$\times$192)*(192$\times$768) and (196$\times$768)*(768$\times$192). 
}
\vskip 0.1in
\begin{tabular}{ 
cccccccc} 

 
 \toprule
  & & \multicolumn{4}{c}{\textbf{FPGA}} & \multicolumn{2}{c}{\textbf{ASIC} \textbf{(45nm)}} \\ 
 \cmidrule(lr){3-6}
 \cmidrule(lr){7-8}
 Method & 1/Xput (cycles) & LUTs & FFs & {BRAMs}  & Energy/ Sample & Area (mm$^2$) & Energy/ Sample\\
\midrule
 8$\times$8   Systolic & 921,600 & 7,792 & 3,178 & 0 & 585.22 uJ & 0.95 & 52.5 uJ \\
 16$\times$16 Systolic & 239,616 & 25,266 & 12,522 & 3  & 327.07 uJ & 1.99 & 28.18 uJ \\
 32$\times$32 Systolic & 64,512 & 94,619 & 50,545 &  11 & 188.69 uJ & 4.31 & 16.12 uJ \\
 64$\times$64 Systolic & 18,432 & 370,939 & 203,023 &  27  &   135.38 uJ      & 10.01 & 10.34 uJ \\
 \textbf{QuWeiT} & 196 & 23,984 & 6,654 & 0   & 0.15 uJ & 0.11 & 0.001 uJ \\
 \bottomrule
\label{table:matmul}
\end{tabular}



\centering
\caption{End-to-end energy consumption comparison between the baseline and QuWeiT for a single encoder layer, per sample inference. A 32x32 systolic array is used for the baseline model and for the non-weightless layers in QuWeiT. We ignore the energy consumption of the other components including SoftMax, GELU and LayerNorm for clarity in analysis, as these were found to not contribute majorly. 
}
\vskip 0.1in
\begin{tabular}{ ccccc}
 \toprule
   & \multicolumn{2}{c}{\textbf{FPGA}} & \multicolumn{2}{c}{\textbf{ASIC} \textbf{(45nm)}} \\ 
    \cmidrule(lr){2-3}
 \cmidrule(lr){4-5}
 Stage & Baseline Model & QuWeiT & Baseline Model & QuWeiT \\
 \midrule
 $Q$, $K$, $V$ & 23.59 uJ (each) & 23.59 uJ (each) & 2.02 uJ (each)& 2.02 uJ (each) \\
 $QK^T$ & 27.52 uJ & 27.52 uJ & 2.35 uJ& 2.35 uJ \\
 $S.V$ & 27.52 uJ & 27.52 uJ & 2.35 uJ & 2.35 uJ \\
 Multi-Head Concat & 23.59 uJ & 23.59 uJ & 2.02 uJ & 2.02 uJ \\
 MLP Dense 1 & 94.35 uJ &  & 8.06 uJ & \\
 MLP Dense 2 & 94.35 uJ & \multirow{-2}{*}{0.15 uJ} & 8.06 uJ & \multirow{-2}{*}{0.001 uJ}\\
 \midrule
 \textbf{Total} & \textbf{338.10 uJ} & \textbf{149.55 uJ} & \textbf{28.90 uJ}& \textbf{12.78 uJ}\\
 \bottomrule
 \vspace{-5mm}
 \label{table:end-to-end}
\end{tabular}

\end{table*}


\subsubsection{Implementation and Energy Evaluation}\label{sec:hwperf}

Considering most implementations on hardware devices are with integer precision, we primarily demonstrate the hardware performance of our solution with the I-ViT-T model. However, it is to be noted that the performance benefits projected with this model hold good for any transformer-based model, as the nature and fraction of computations is similar across these, as indicated in Section \ref{sec:intro}.

First, we make a direct comparison of the performance of the weightless block's PE as compared to a systolic array based implementation of the MLP block that is being replaced. For the I-ViT-T model, the MLP layer involves two GEMMs of dimensions 1) $196 \times 192$ by $192 \times 768$ and 2) $192 \times 768$ by $768 \times 192$, with an intermediate ShiftGELU activation layer. The ShiftGELU, proposed in \cite{ivit} is a hardware-optimized version of the conventional GELU operation. As described in Section \ref{sec:eval}, we project the latency (1/throughput) and energy consumption on the target FPGA and ASIC for the GEMMs under consideration. 

Table \ref{table:matmul} summarizes these findings. As shown, the PE for weightless block outperforms different systolic array based baseline implementations of the corresponding MLP layer, consuming only 0.15uJ energy per sample on the FPGA - a $900 \times$ improvement over the best case systolic array. Furthermore, with QuWeiT, we also completely eliminate any BRAMs required for weights or partial product staging. The performance improvement on a custom ASIC is even more staggering, with only 0.001uJ energy consumed per sample - a roughly $10000 \times$ improvement. We note that as we increase the systolic array dimensions, though the energy required for processing a sample through the MLP decreases, the resource consumption for the systolic array block increases significantly. 



In order to provide a more holistic evaluation, we compare the end-to-end consumption per sample inference through a single model encoder layer 
for the baseline and QuWeiT models, based on a streaming architecture as described in Section \ref{sec:eval}. 
We considered the 32x32 systolic array architecture for this comparison for an optimal area-energy tradeoff. Note that since we are using the same architecture as the original model for the parts of QuWeiT that remain unchanged, the energy associated with the Multi-head self-attention (MSA) layers remains unchanged. Table \ref{table:end-to-end} summarizes these results, and shows the QuWeiT implementation outperforming the baseline model implementation by a significant degree. We note that the layers with the highest energy consumption (MLP Dense 1 and MLP Dense 2) are the ones replaced with the weightless block, and hence provides a significant reduction in energy consumption. 
The total end-to-end energy consumption for a single sample with the baseline model implementation indicates over $2.2\times$ improvement in energy efficiency for both the target FPGA and ASIC. A similar analysis done for the nanoGPT based network shows $2.5\times$ savings in energy. 

With the weightless block implementation, we are completely eliminating about 66\% of the overall model weights. We note that typically a lot of these model weights for MLP layers are stored off-chip and are fetched on-demand. Therefore, with QuWeiT we also eliminate the corresponding off-chip memory accesses. The resultant savings in reduced weight movement to on-chip memory would be much more pronounced.
These results on performance and power consumption lend support to the immense potential of QuWeiT models over conventional transformers.

\vspace{-1mm}
\section{Discussion} \label{sec:discuss}

While we illustrate performance benefits only for IViT and nanoGPT, due to the structural similarity of all transformers, similar energy savings are expected in any large networks, including GPT-3. Due to the large training times and resources required to train a model of the class of GPT3, we restricted ourselves to smaller networks. However, the model performance trends across vision and language tasks demonstrate the versatility of the solution.
In the current context of energy-efficient inference deployment solutions, the presented evaluation still proves Quasi-Weightless Transformers to be a competent alternative to conventional transformer-based models. Furthermore, the demonstrated tasks are definitely relevant for edge applications, where energy-critical solutions are more important to meet the limited on-device energy targets. We also view these models as a stepping stone towards fully-weightless transformer architecture design, and envision a class of tiny language models that are competitive to the current energy-inefficient LLMs. 

A key limitation that we note of this work, is the relatively long training time for these QuWeiTs.
Training the weightless block we introduce into the network involves several bit-manipulation operations, which are less optimized on GPUs compared to FLOPs.
However, since this work is primarily designed as an inference-efficient solution, the one-time training cost, which is amortized over numerous inferences after deployment, is not a major concern.

Another aspect to note is that in hardware accelerators, the systolic array block could be reused to perform various matrix multiplications across different layers in the model. On the other hand, the PE block is specific for the particular weightless layer, as the learned encodings are embedded in the LUT or gate-level implementation on the FPGA or ASIC.  However, this is not a cause for concern when deploying on a large FPGA, where the operations across layers are streamed through in a pipeline fashion through different compute units.

\vspace{-2mm}
\section{Related Work} \label{sec:related}

In recent years, numerous studies have explored energy-efficient transformer architectures and their hardware accelerator designs. SwifTron \cite{swiftron} introduces a specialized integer-arithmetic-based accelerator tailored for transformers, optimizing both energy efficiency and computational speed, but retains the original model architecture as is.
Model quantization has been extended to extreme levels to enable memory and compute-efficient, multiplication-free models. BinaryBERT \cite{binary-bert} and Bitnet 1.58b \cite{bitnet1} advance this concept by employing 1-bit quantization, where traditional multiply-accumulate operations are replaced with ternary or binary sum operations. This approach enables these models to achieve performance comparable to full-precision counterparts, with higher energy efficiency. However, these techniques do not reduce the model parameters. 

Recently, there has been a surge of interest in approximate matrix multiplication (AMM) to enhance energy efficiency. 
A transformer architecture was proposed that approximates floating-point multiplications with piece-wise affine functions, achieving comparable performance while reducing computational demands primarily for energy savings \cite{affine}. 
LUT-GEMM \cite{lut-gemm-iclr} introduces an AMM method using LUTs to reduce both energy consumption and latency by modifying GPU kernels.
However, these methods primarily rely on LUTs as an inference-time optimization, substituting multiplication operations with lookup operations to achieve post-training speedup. In contrast, our work focuses on WNNs, a fundamentally different approach that employs LUTs as the core computational units or ``neurons" of the model, learning them directly during training. This integration of LUTs at the neural architecture level goes beyond inference optimizations, aiming to fully utilize LUTs learning capacity in the training process for improved efficiency. We note that our technique is infact complimentary to works like Bitnet \cite{bitnet1}, and a quasi-weightless transformer based design can be integrated with a binarized transformer to further enhance the model-size and energy reduction achieved by those models.

\vspace{-2mm}
\section{Conclusion} \label{sec:conclusion}
In this work we introduce Quasi-Weightless Transformers, an effort to shrink the giant transformers to achieve a low energy inference alternative. We identify an opportunity to increase the energy-efficiency of these model inferences by targeting the most compute and memory intensive layers in transformers and replacing them with Differentiable Weightless Neural Network layers. By eliminating the MLP layers in conventional transformer-based models and introducing a weightless block, we achieve an overall 50-70\% reduction in computations and model weights across typical transformer models. 
Based on our model and performance evaluations, we report significant energy savings when deployed on FPGAs and ASICs across vision and language tasks, with similar model performance. These results illustrate that Quasi Weightless Transformers are a promising lightweight alternative to traditional transformers - paving the way for tiny language models. While WNNs have been applied to many small-scale problems in the past, this is the first time that its usefulness in shrinking a giant transformer model is demonstrated.

\section*{Acknowledgements}
This research was supported in part by the Semiconductor Research Corporation (SRC) Task 3148.001, NSF Grant \#2326894, and NVIDIA Applied Research Accelerator Program Grant. 

\bibliographystyle{mlsys2024}
\bibliography{mlsys_paper}


\appendix

\end{document}